\documentclass{article}

\usepackage[final]{nips_2018}

\usepackage[utf8]{inputenc} 
\usepackage[T1]{fontenc}    
\usepackage{hyperref}       
\usepackage{url}            
\usepackage{booktabs}       
\usepackage{amsfonts}       
\usepackage{nicefrac}       
\usepackage{microtype}      
\usepackage{float}
\usepackage[final]{graphicx}
\usepackage{subcaption}
\usepackage{multirow}
\usepackage{booktabs}
\usepackage{tablefootnote}
\usepackage{authblk}

\title{Deep Learning Approach for Predicting 30 Day Readmissions after Coronary Artery Bypass Graft Surgery}

\author[1,2]{Ramesh B.~Manyam}
\author[1]{Yanqing Zhang}
\author[2]{William B.~Keeling}
\author[2]{Jose Binongo}
\author[2]{Michael Kayatta}
\author[3]{Seth Carter}
\affil[1]{Georgia State University}
\affil[2]{Emory University}
\affil[3]{University of North Georgia}

\begin{document}

\maketitle

\begin{abstract}
Hospital Readmissions within 30 days after discharge following Coronary Artery Bypass Graft (CABG) Surgery are substantial contributors to healthcare costs. Many predictive models were developed to identify risk factors for readmissions. However, majority of the existing models use statistical analysis techniques with data available at discharge. We propose an ensembled model to predict CABG readmissions using pre-discharge perioperative data and machine learning survival analysis techniques. Firstly, we applied fifty one potential readmission risk variables to Cox Proportional Hazard (CPH) survival regression univariate analysis. Fourteen of them turned out to be significant (with p value < 0.05), contributing to readmissions. Subsequently, we applied these 14 predictors to multivariate CPH model and Deep Learning Neural Network (NN) representation of the CPH model, DeepSurv. We validated this new ensembled model with 453 isolated adult CABG cases. Nine of the fourteen perioperative risk variables were identified as the most significant with Hazard Ratios (HR) of greater than 1.0. The concordance index metrics for CPH, DeepSurv, and ensembled models were then evaluated with training and validation datasets. Our ensembled model yielded promising results in terms of c-statistics, as we raised the the number of iterations and data set sizes. 30 day all-cause readmissions among isolated CABG patients can be predicted more effectively with perioperative pre-discharge data, using machine learning survival analysis techniques. Prediction accuracy levels could be improved further with deep learning algorithms. 

\end{abstract}

\section{Introduction}

Healthcare costs are constantly rising around the globe with each passing year. Hospital readmissions, that occur shortly after discharge, are identified as substantial contributors to these escalating healthcare expenses. So, in an effort to reduce readmissions, Affordable Care Act established ‘Hospital Readmission Reduction Program’ [HRRP] in 2012 [1]. This program imposes financial penalties to hospitals that have higher rates of readmissions. For FY 2018, the HRRP added six measures of medical conditions, and one of them is Coronary Artery Bypass Graft surgery (CABG). Average medicare readmission rate for CABG patients is 1 in 5 [2, 3]. In many situations, these readmissions are potentially avoidable. Therefore, the primary focus of our research study is to develop a comprehensive predictive model that identifies the most significant risk factors associated with 30 day readmissions following CABG surgery, using multivariate survival regression analysis techniques and deep learning neural network algorithms as well. Specifically, our study focuses on applying i) patients demographics data and perioperative characteristics, ii) time-varying pre-discharge clinical data as well, to an enhanced Cox Proportional Hazards (CPH) [4] model. one of the popular logistic regression methods used in Survival Analysis problem solving areas [22]. Also, our model considers pre-discharge time-varying clinical data and applies a new ensembled approach - feeding only the relevant patient data to survival regression analysis, getting the highly-correlated significant risk factors contributing to 30 day readmissions after discharge, applying the study cohort to cross validation, and eventually to a customized deep survival neural network, DeepSurv [17].

\section{Related Work}
Currently, most of the studies on cardiac surgery outcomes used third-party statistical tools for data analysis, such as SAS and STATA software tools [6-9]. These techniques and their prediction accuracy levels vary from hospital to hospital, disease to disease, database to database. Majority of the studies use traditional t-tests, chi-square, Fisher exact, Wilcoxon rank sum tests, stepwise logistic regression, Hosmer-Lemeshow goodness-of-fit tests, dealing with categorical and continuous variables for univariate and multivariate analyses [5-11].  Fanari et al. used Hierarchical logistic regression to model readmissions and Fractional polynomial (FP) regression to assess non-linearity of continuous variables [14]. Also, most of the current predictive analytic models for 30 day CABG readmissions consider a single-point, time invariant patient characteristics available at discharge, while Benuzillo et al. evaluated readmission risk with patient data available shortly after admission [5]. However, pre-discharge time-varying labs, vitals, medications, post-discharge follow-up care measures do play a significant role in determining risk factors for hospital  readmissions  [15]. Recently, Katzman et al. developed and used a deep Cox proportional hazards neural network, also known as, DeepSurv [16]. DeepSurv model outperformed regular CPH on survival data with both linear and non-linear risk functions [17]. Some more survival analysis models that dealt with healthcare applications include deep active analysis, deep recurrent analysis, deep integrative analysis, transfer learning as well [18 - 21]. 

\begin{figure}
  \centering
  \includegraphics[width=1.0\linewidth]{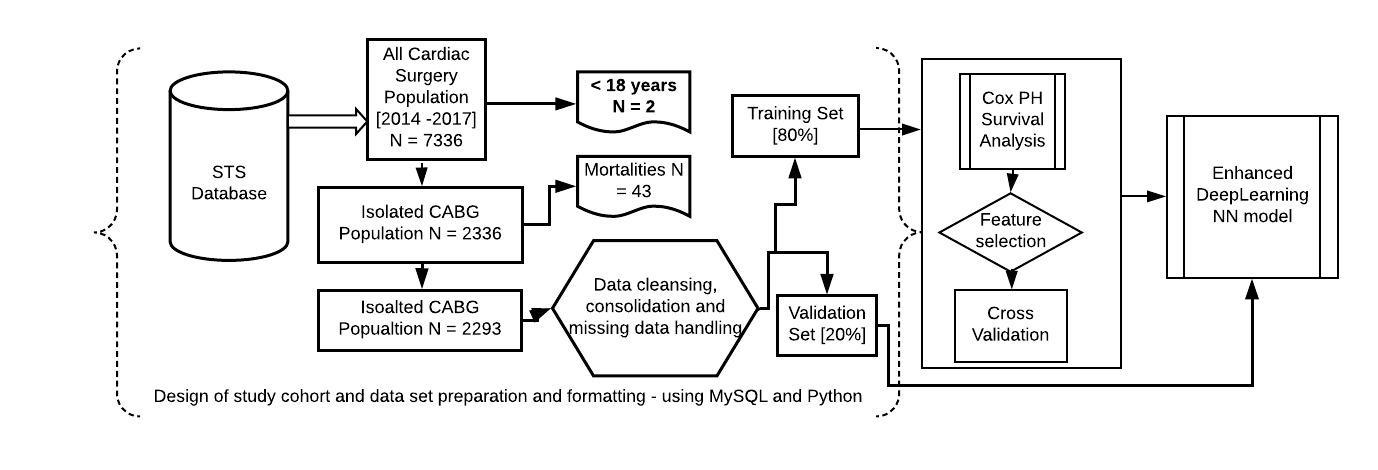}
  \caption{Block diagram of the proposed CABG readmission prediction framework}
\end{figure}

\section{Proposed Model} 
In this section, our proposed framework and its implementation would be described. The proposed model is shown in Figure 1. Our model is an ensemble of two models, regular CPH as well as DeepSurv [16]. Firstly, we prepared the study population cohort through rigorous data cleansing and consolidation process and fed it to the CPH regression analysis in an appropriate format [22], shortlisted the most significant covariates, and then applied the resultant normalized data set to an open source DeepSurv python package [23]. We implemented our model through customizing open source Python packages, such as Lifelines and Deepsurv. Lifelines   is an implementation of survival analysis and DeepSurv is a deep learning version of the Cox proportional hazards model that uses Theano and Lasagne [22- 23] python packages as well. DeepSurv has an advantage over traditional Cox regression because it does not require an apriori selection of covariates, but learns them adaptively.

\begin{table}
\caption{Characteristics of significant risk predictors from CPH survival regression model}
\label{sample-table-1}
\centering
\begin{tabular}{lllllll}
\toprule 
    Variable & \multicolumn{3}{c}{Univariate Analysis} & \multicolumn{3}{c}{Multivariate Analysis}\\

    & $\beta$ coeff. & HR\footnotemark[1]  & p value
    & $\beta$ coeff. & HR\footnotemark[1] &  p value  \\
    \midrule
     Gender     & 0.3554   &1.4268  &<0.05 &0.1168 &1.1239 &0.4463\\
    
    Alcohol Use     &0.1164   &1.1235  &<0.05 &0.0750 &1.0779 &0.1586 \\
    Cerebrovascular Disease   & -0.4107  &0.6632    & <0.05 & -0.1984 &0.8200 &0.1822\\
    
    Chronic Lung Disease     & 0.1294    &1.1382    &<0.05 &0.0574 &1.0590 &0.3074 \\
      Preoperative Creatinine Level    & 0.1205    &1.1281    &  <0.001 & 0.0327 & 1.0333 &0.7433\\
    
    Preoperative Hematocrit     & -0.0480    &0.9531    &  <0.001 &-0.0263 & 0.9741 &0.0282 \\
    Total Bilirubin     &-0.4729    &0.6831    &  <0.05 &-0.3455 & 0.7079 &0.0844\\
    
    Prior Myocardial Infarction     &0.0171    &0.6632    &  <0.05 & 0.0264 & 1.0267 &0.7666 \\
      Intraoperative Blood Products      & -0.4107    &0.6917    & <0.05 & 0.0630 & 1.0650 &0.6888\\
    
    Postoperative Blood Products     &-0.5995    &0.5491    &  <0.001 &-0.3147 & 0.7300 &0.0368 \\
    Postoperative Creatinine Level     & 0.1258    &1.1341    &  <0.001 & 0.0264 & 1.0267 &0.7666\\
    
   Total ICU Hours     & 0.0021    &1.0021    &  <0.001 & 0.0007 & 1.0007 &0.4438 \\
      Postoperative Events     & -0.2835    &0.7532    &  <0.05 & -0.0610 & 0.9409 &0.6773\\
    
    Length of Stay    & 0.0340   &1.0346   &  <0.001 & 0.0089 & 1.0089 &0.5179 \\
       
\bottomrule
\end{tabular}

\end{table}
\footnotetext[1]{Hazard Ratio.}

\begin{table}
\caption{Concordance index measures for training and validation data sets from 3 different models, Cox PH, DeepSurv, and proposed ensembled model, with 95\% confidence interval for the mean C-index obtained using the same set of study cohorts}
\label{sample-table-2}
\centering
\begin{tabular}{llllll}
\toprule 
    Dataset & CoxPH & DeepSurv &\multicolumn{3}{c}{Ensembled Model}\\

    & model & nEpochs=10000  & nEpochs=10000  & nEpochs=20000 & nEpochs=25000 \\
    \midrule
     Training     & 0.660   &0.6732  &0.665 &0.696 &\textbf{0.712} \\
    
    Validation     &0.631   &0.535 &0.546 &0.569 &0.571  \\

\bottomrule
\end{tabular}
\end{table}

\section{Experiments and Results} 
As shown in Figure 1,  the primary database used in this study is Society of Thoracic Surgeons (STS) Adult Cardiac Surgery Database, Version 2.81, pertaining to a single U.S institution, multi-hospital facility. A de-identified data set of adult cardiac surgery population for the period of July 2014 to May 2017 were extracted, and processed with Structured Query Language (SQL) scripts, with the aid of an open source MySQL workbench software [24]. Excluding patients of < 18 years of age, and in-hospital mortalities, a cleaned, and consolidated data set of 2293 adult CABG cases surviving to discharge are the subjects of this preliminary study. We used single imputation techniques, such as, case deletion and conditional mean and distribution for handling missing data in this iteration. Then, we divided the dataset randomly into training and validation sets of 80:20, with 51 potential variables representing study population’s demographics data (such as age, race, sex etc.), and perioperative characteristics. The primary outcome was all-cause readmission within 30 days of hospital discharge. The training data set was first fed into univariate analysis, using CPH package of Lifelines [22], and then the 14 most significant variables with a p value of <0.05 are considered for multivariate analysis. The corresponding statistics, comprising of $\beta$ coefficients, hazard ratios,  and p-values, are shown in Table 1. The first column in the table represents an explanatory variable or predictor, and the results of univariate and multivariate analyses are shown in the subsequent columns. For instance, for the variable, ‘Gender’, the exponential coefficient (exp(coef) = exp(0.3554) = 1.4268), also known as Hazard Ratio(HR), implies the effect size of covariates. It can be inferred that, a female gender would have hazard of readmission by a factor of 1.4268, compared to a male. With fitted model, we presented 95 percent Confidence Interval(CI) ranges of a few significant variables and survival plots, as shown in Figure 2. Concordance index statistics, that represent a generalization of the area under the Receiver Operating Characteristics curve, are obtained through DeepSurv Neural Network model, and a couple of plots are depicted in Figure 3. Table 2 shows the comparison of concordance metrics for our initial set of training and validation cohorts and it is evident that the ensembled model yields better performance with increasing number of iterations, for instance, c-index of 0.712 with nEpochs = 25000. We anticipate the prediction accuracy levels of training as well as validation sets to be improved substantially, when we feed huge amounts of pre-discharge data in our next iteration.

\section{Conclusion and future work}
30-day all‐cause readmissions among isolated CABG patients can be predicted prior to discharge, more effectively with perioperative pre-discharge data, using machine learning survival analysis techniques. Prediction accuracy levels could be improved further with deep learning algorithms, feeding huge amounts of highly-correlated pre-discharge data, including time-dependent lab values, vitals, and medications. With our ensemble modeling approach, concordance statistics for training and validation sets have shown considerable improvements with increasing nEpochs and volume of data, and averaged slightly above 0.65 for the initial iterations. With this model, feeding enormous amounts of pre-discharge data, we intend to develop a Risk of Readmission (RoR) score, for patient population undergoing CABG surgery. An additional aim of this study is to create a smart-app for bedside use by clinicians to enable them to more appropriately decide patient readiness for discharge. 

\begin{figure}[h]
 

\begin{subfigure}{0.34\textwidth}
  \centering
  \includegraphics[width=\linewidth]{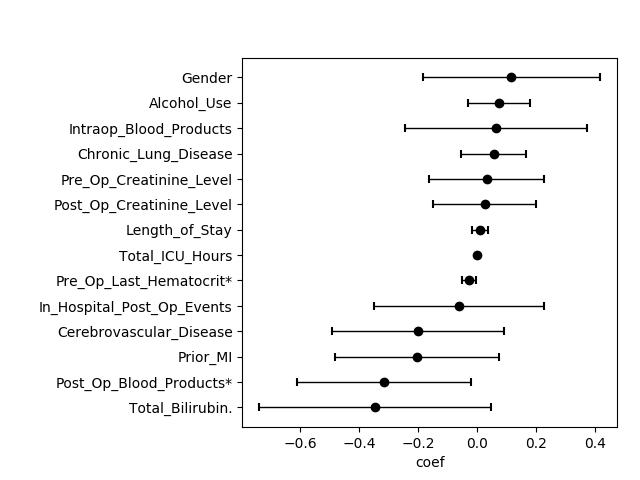}%
\end{subfigure}\begin{subfigure}{0.32\textwidth}
  \centering
  \includegraphics[width=\linewidth]{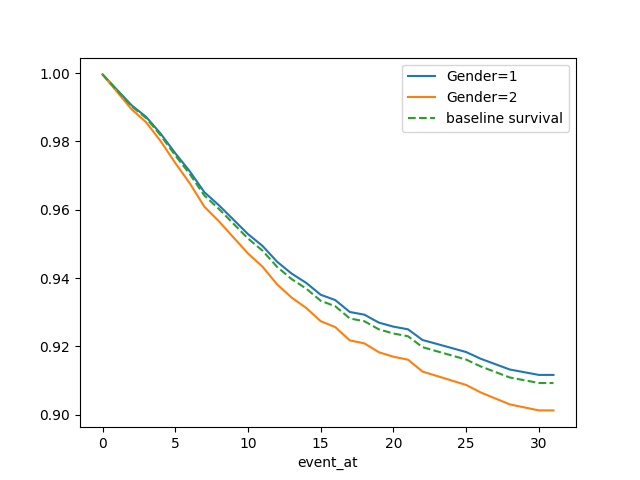}%
\end{subfigure}
\begin{subfigure}{0.32\textwidth}
  \centering
  \includegraphics[width=\linewidth]{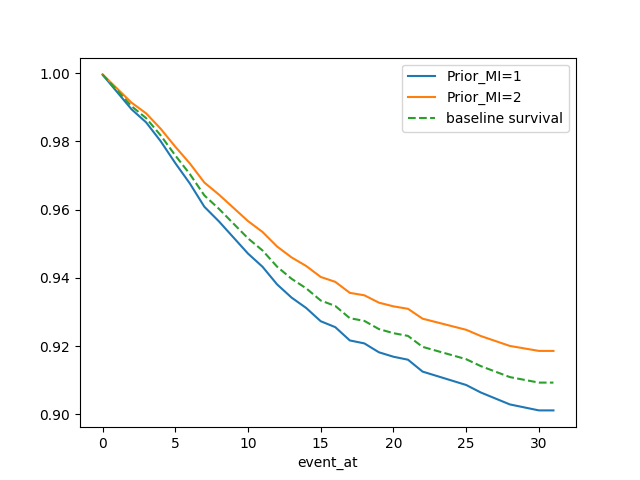}%
\end{subfigure}\\
\begin{subfigure}{0.34\textwidth}
  \centering
  \includegraphics[width=\linewidth]{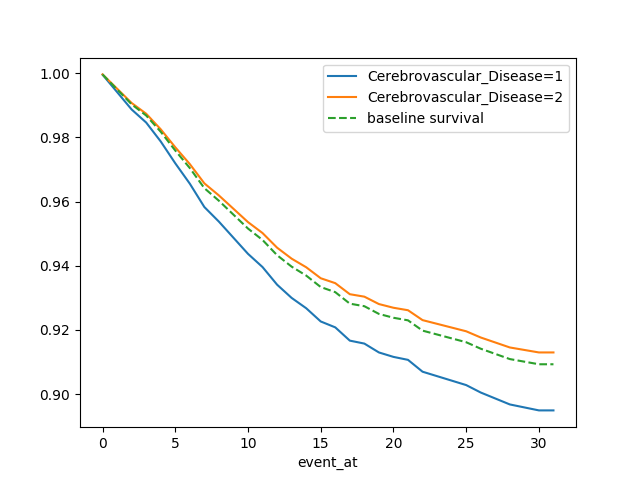}%
\end{subfigure}\begin{subfigure}{0.32\textwidth}
  \centering
  \includegraphics[width=\linewidth]{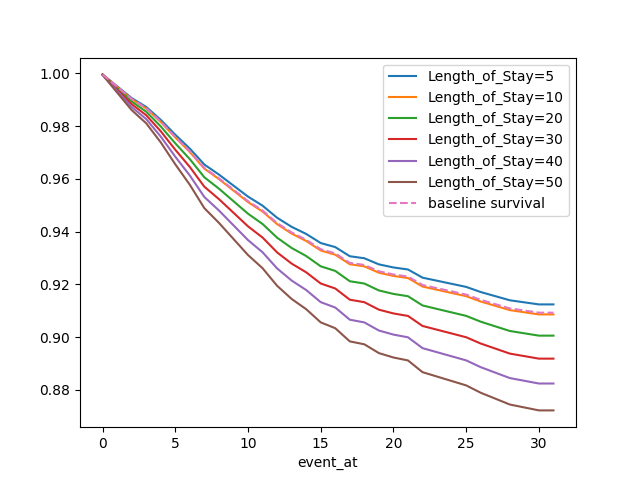}%
\end{subfigure}
\begin{subfigure}{0.32\textwidth}
  \centering
  \includegraphics[width=\linewidth]{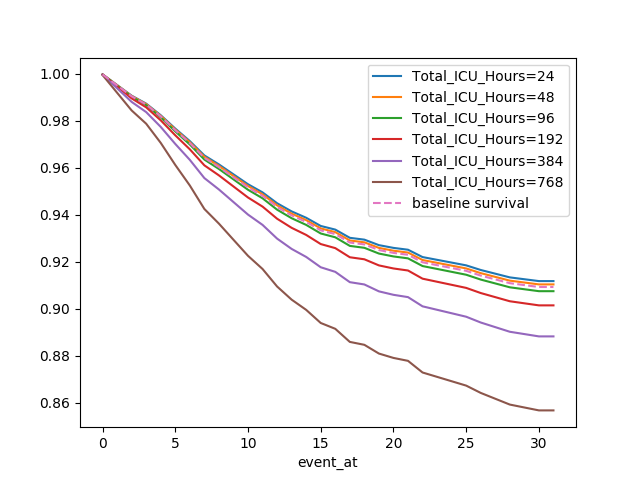}%
\end{subfigure}\\
\caption{95 percent Confidence Interval(CI) ranges and survival curves from Cox PH model multivariate analysis for a few significant risk predictor variables, such as, Gender, 1=Male, 2=Female; Prior Myocardial Infarction(Prior MI), 1= Yes, 2= No; Cerebrovascular disease, Length of Stay, and Total ICU hours spent in hospital}

\begin{subfigure}{0.34\textwidth}
  \centering
  \includegraphics[width=\linewidth]{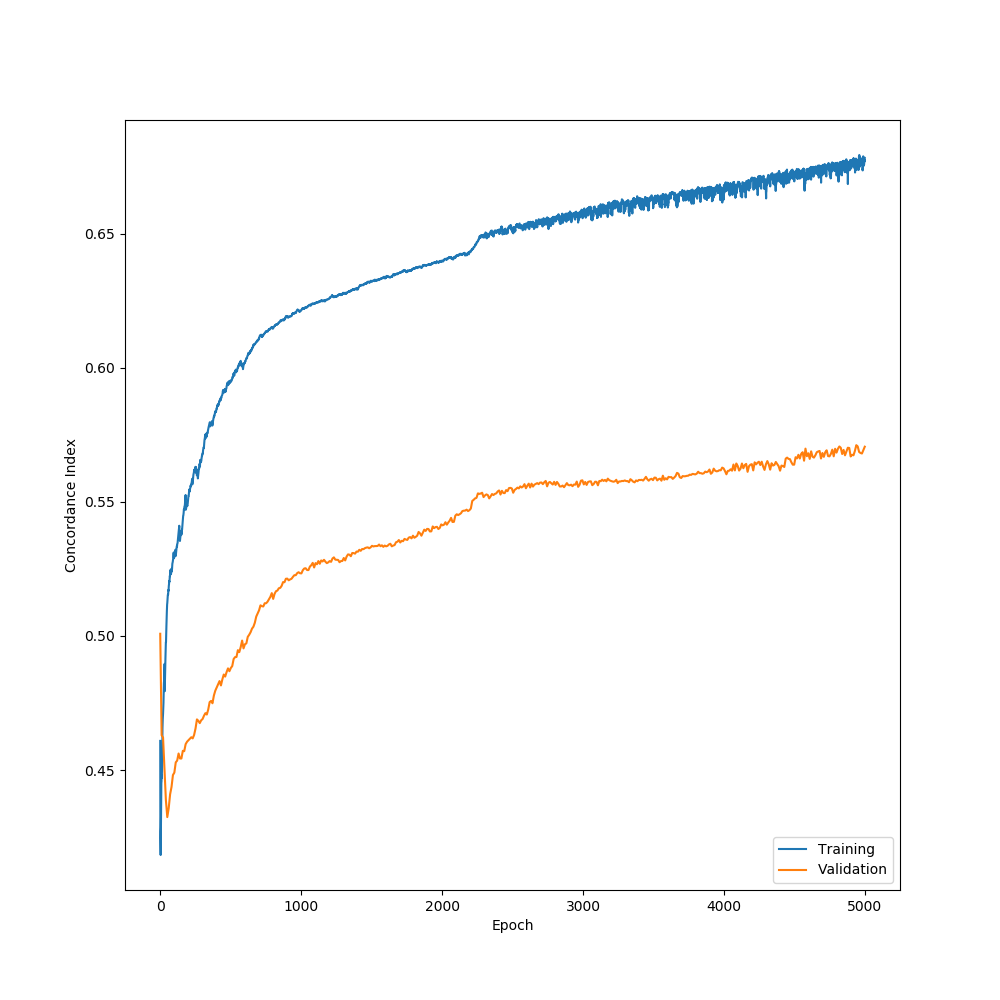}%
\end{subfigure}
\begin{subfigure}{0.32\textwidth}
  \centering
  \includegraphics[width=\linewidth]{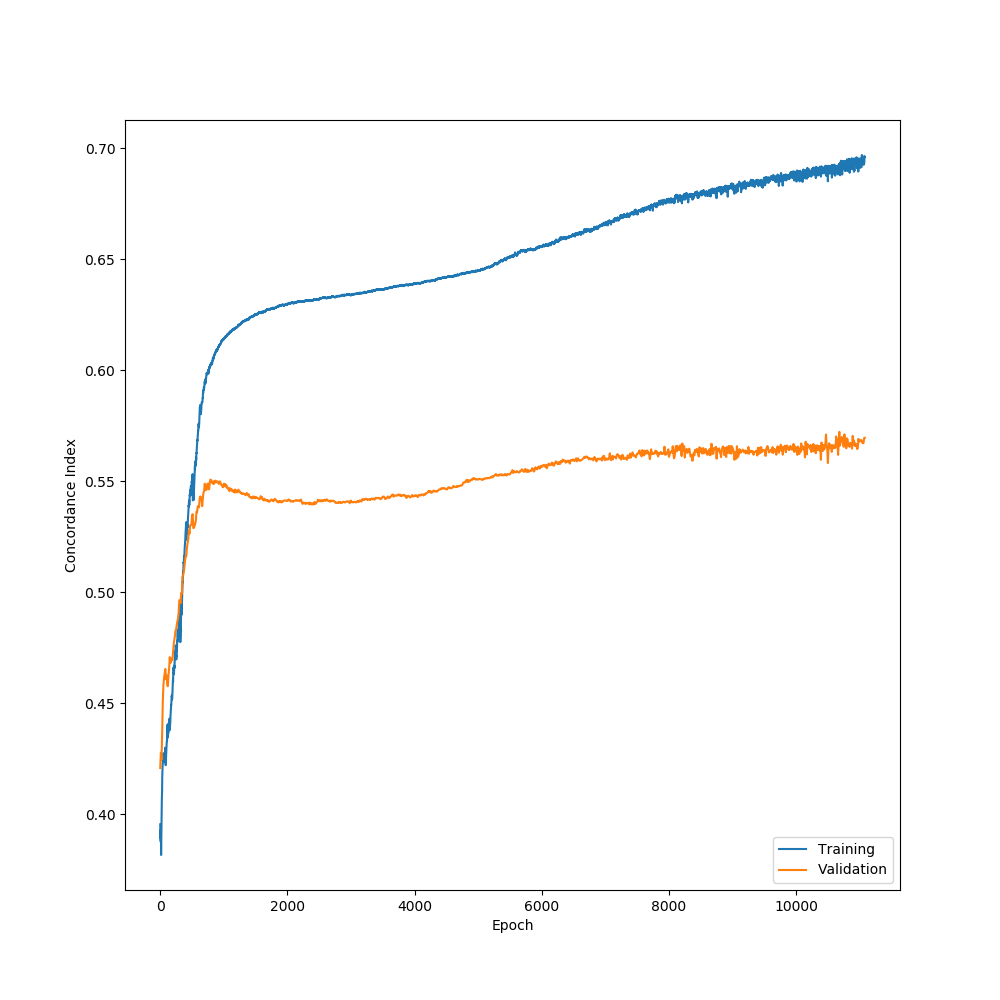}%
\end{subfigure}
 \begin{subfigure}{0.32\textwidth}
  \centering
  \includegraphics[width=\linewidth]{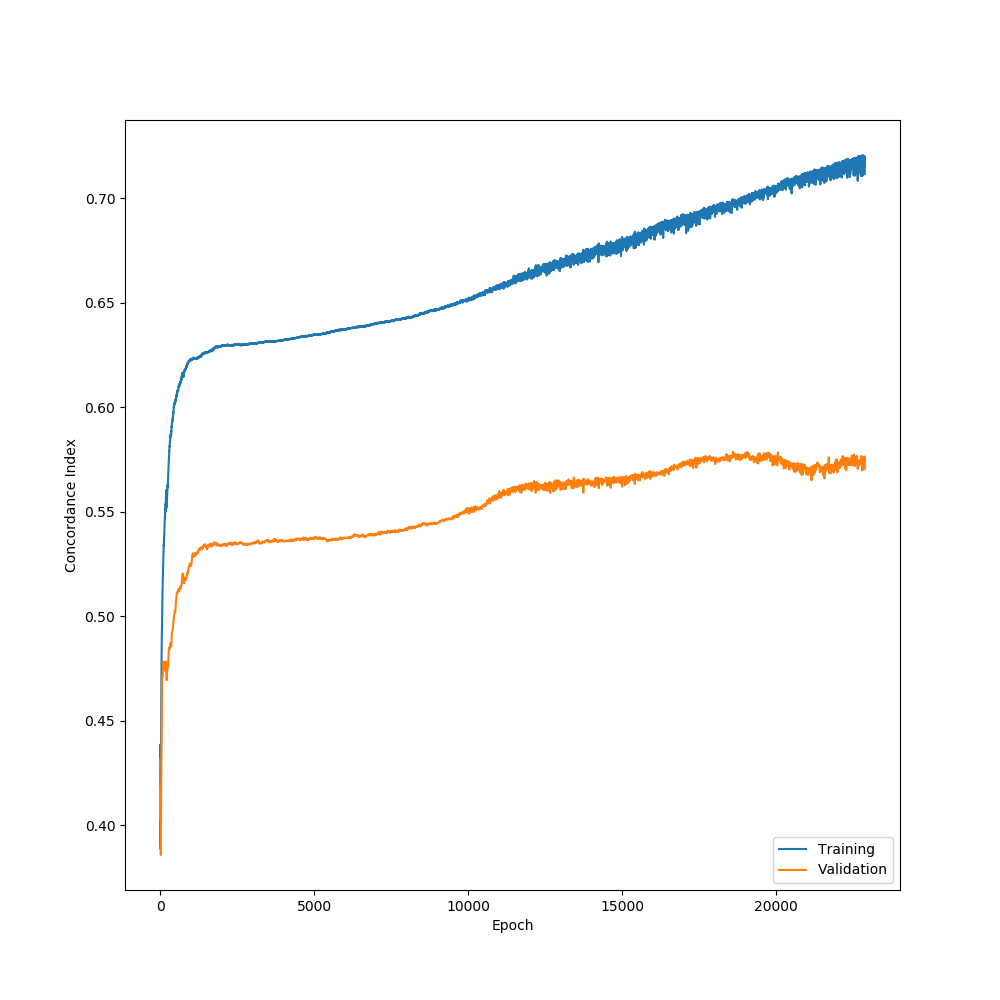}%
\end{subfigure}
\caption{Concordance index measures for training and validation datasets from proposed model}

\end{figure}

\section*{References}
\medskip

\small



[1] Centers for Medicare and Medicaid Services.Readmissions Reduction Program. Available at: https://www.cms.gov/medicare/medicare-fee-for-service-payment/acuteinpatientpps/readmissions-reduction-program.html. Accessed December 20, 2017.  

[2] Jencks SF, Williams MV, \& Coleman EA (2009) Rehospitalizations among patients in the Medicare fee-for-service program. New England Journal of Medicine 360 (14):1418-1428.

[3] Osnabrugge RL, Speir AM, \& Head SJ, et al. Cost, quality, and value in coronary artery bypass grafting. The Journal of Thoracic and Cardiovascular Surgery. 2014;148(6):2729–2735.e1. doi:10.1016/j.jtcvs.2014.07.089.

[4] David R Cox. Regression models and life-tables. In Breakthroughs in statistics. Springer, 1992.

[5] Benuzillo, J., Caine, W., Evans, R., Roberts, C., Lappe, D., \& Doty, J. (n.d.). Predicting readmission risk shortly after admission for CABG surgery. Journal of Cardiac Surgery., 33(4), 163-170.

[6] Feng, T., White, R., Gaber-Baylis, L., Turnbull, Z., \& Rong, L. (n.d.). Coronary artery bypass graft readmission rates and risk factors - A retrospective cohort study. International Journal of Surgery., 54(Pt A), 7-17.

[7] Guru, V., Fremes, S., Austin, P., Blackstone, E., \& Tu, J. (n.d.). Gender differences in outcomes after hospital discharge from coronary artery bypass grafting. Circulation : Journal of the American Heart Association., 113(4), 507-516.

[8] Stewart, R., Campos, C., Jennings, B., Lollis, S., Levitsky, S., \& Lahey, S. (n.d.). Predictors of 30-day hospital readmission after coronary artery bypass. The Annals of Thoracic Surgery., 70(1), 169-174.

[9] E.L. Hannan, Y. Zhong, S.J. Lahey, A.T. Culliford, J.P. Gold, C.R. Smith, R.S.D. Higgins, D. Jordan, \& A. Wechsler, 30-day readmissions after coronary artery bypass graft surgery in New York State, JACC Cardiovasc. Interv. 4 (2011) 569–576, http://dx.doi.org/10.1016/j.jcin.2011.01.010.

[10] Maniar, H., Bell, J., Moon, M., Meyers, B., Marsala, J., Lawton, J., \& Damiano, R. (n.d.). Prospective evaluation of patients readmitted after cardiac surgery: Analysis of outcomes and identification of risk factors. The Journal of Thoracic and Cardiovascular Surgery., 147(3), 1013-1018.

[11] Kilic, Magruder, Grimm, Dungan, Crawford, Whitman, \& Conte. (2017). Development and Validation of a Score to Predict the Risk of Readmission After Adult Cardiac Operations. The Annals of Thoracic Surgery, 103(1), 66-73.

[12] Price, J., Romeiser, J., Gnerre, J., Shroyer, A., \& Rosengart, T. (n.d.). Risk analysis for readmission after coronary artery bypass surgery: Developing a strategy to reduce readmissions. Journal of the American College of Surgeons., 216(3), 412-419.

[13] Vaccarino, V., Lin, Z., Kasl, S., Mattera, J., Roumanis, S., Abramson, J., \& Krumholz, H. (2003). Sex differences in health status after coronary artery bypass surgery. Circulation : Journal of the American Heart Association., 108(21), 2642-2647.

[14] Fanari, Z., Elliott, D., Russo, C., Kolm, P., \& Weintraub, W. (n.d.). Predicting readmission risk following coronary artery bypass surgery at the time of admission. Cardiovascular Revascularization Medicine., 18(2), 95-99.

[15] Gao, D., Grunwald, G., Rumsfeld, J., Schooley, L., MacKenzie, T., \& Shroyer, A. (n.d.). Time-varying risk factors for long-term mortality after coronary artery bypass graft surgery. The Annals of Thoracic Surgery., 81(3), 793-799.

[16] Katzman, J., Shaham, U., Cloninger, A., Bates, J., Jiang, T., \& Kluger, Y.: Deep survival: A deep Cox proportional hazards network. CoRR (2016)

[17] Katzman, J., Shaham, U., Cloninger, A., Bates, J., Jiang, T., \& Kluger, Y. (2018). DeepSurv: Personalized treatment recommender system using a Cox proportional hazards deep neural network. BMC Medical Research Methodology, 18(1), 24.

[18] Nezhad, Sadati, Yang, \& Zhu. (2019). A Deep Active Survival Analysis approach for precision treatment recommendations: Application of prostate cancer. Expert Systems With Applications, 115, 16-26.

[19] Giunchiglia, Eleonora \& Nemchenko, Anton \& van der Schaar, Mihaela. (2018). RNN-SURV: a Deep Recurrent Model for Survival Analysis. 

[20] Huang, C., Zhang, A., \& Xiao, G. (2018). Deep Integrative Analysis for Survival Prediction. Pac Symp Biocomput, 23, 343-352.

[21] Li, Yan, Wang, Lu, Wang, Jie, Ye, Jieping, \& Reddy, Chandan K. (2016). Transfer Learning for Survival Analysis via Efficient L2,1-Norm Regularized Cox Regression. Data Mining (ICDM), 2016 IEEE 16th International Conference on, 231-240.

[22] Davidson-Pilon C. Lifelines. https://github.com/camdavidsonpilon/lifelines, 2016.

[23] Katzman J., DeepSurv, https://github.com/jaredleekatzman/DeepSurv, 2016.

[24] MySQL Workbench, Open Source Software, https://dev.mysql.com/doc/workbench/en/, 2018 

\end{document}